\documentclass[a4paper,11pt]{article}

\usepackage[scaled=1.0]{helvet}

\usepackage[T1]{fontenc}
\usepackage[utf8]{inputenc}

\usepackage[a4paper,
            top=2.54cm,
            bottom=2.54cm,
            left=2.54cm,
            right=2.54cm,
            headheight=50pt]{geometry}

\usepackage{lineno}
\modulolinenumbers[1]

\usepackage{parskip}
\usepackage{titlesec}

\titleformat{\section}
  {\fontsize{14}{16.8}\selectfont\bfseries\sffamily}
  {\thesection.}{0.5em}{}
\titlespacing*{\section}{0pt}{12pt}{6pt}

\titleformat{\subsection}
  {\fontsize{12}{14.4}\selectfont\bfseries\sffamily}
  {\thesubsection}{0.5em}{}
\titlespacing*{\subsection}{0pt}{6pt}{6pt}

\titleformat{\subsubsection}
  {\fontsize{11}{13.2}\selectfont\bfseries\itshape\sffamily}
  {\thesubsubsection}{0.5em}{}
\titlespacing*{\subsubsection}{0pt}{6pt}{6pt}

\titleformat{\paragraph}[runin]
  {\fontsize{11}{13.2}\selectfont\bfseries\itshape\sffamily}
  {\theparagraph}{0.5em}{}[.\quad]
\titlespacing*{\paragraph}{0pt}{6pt}{0pt}

\usepackage{caption}
\usepackage{xcolor}
\definecolor{atrf@grey}{gray}{0.50}
\definecolor{navy}{rgb}{0.1, 0.1, 0.8}
\definecolor{mygray}{rgb}{0.6, 0.6, 0.6}
\definecolor{myblue}{rgb}{.8, .8, 1}
\definecolor{olive}{rgb}{0.1, 0.5, 0.1}
\definecolor{mymagenta}{rgb}{0.55, 0.0, 0.55}

\usepackage{fancyhdr}

\fancypagestyle{firstpage}{%
  \fancyhf{}
  
  \fancyhead[C]{%
    \fontsize{10}{12}\selectfont\sffamily\color{atrf@grey}%
    \begin{tabular}[b]{c}
      Australasian Transport Research Forum 2026 Proceedings\\[1pt]
      24--26 November, Sydney, Australia\\[1pt]
      Publication website: \textcolor{atrf@grey}{\url{https://australasiantransportresearchforum.org.au/}}%
    \end{tabular}%
  }
  \fancyfoot[R]{\fontsize{10}{12}\selectfont\sffamily\thepage}
}

\usepackage{graphicx}
\graphicspath{{./}}
\DeclareGraphicsExtensions{.eps,.pdf,.png,.jpg,.tif,.tiff,.ps}
\usepackage{amsmath}
\usepackage{amssymb}
\usepackage{amsfonts}
\usepackage{booktabs}
\usepackage{array}
\usepackage{subcaption}
\usepackage{xspace}
\usepackage{url}

\usepackage{natbib}
\usepackage[hidelinks]{hyperref}
\usepackage[nameinlink,capitalize]{cleveref}% cleveref must come after hyperref AND natbib
\AtBeginDocument{\def\harvardurl#1{\url{#1}}}

\newcommand*\patchAmsMathEnvironmentForLineno[1]{%
  \expandafter\let\csname old#1\expandafter\endcsname\csname #1\endcsname
  \expandafter\let\csname oldend#1\expandafter\endcsname\csname end#1\endcsname
  \renewenvironment{#1}%
    {\linenomath\csname old#1\endcsname}%
    {\csname oldend#1\endcsname\endlinenomath}%
}
\newcommand*\patchBothAmsMathEnvironmentsForLineno[1]{%
  \patchAmsMathEnvironmentForLineno{#1}%
  \patchAmsMathEnvironmentForLineno{#1*}%
}
\AtBeginDocument{%
  \patchBothAmsMathEnvironmentsForLineno{equation}%
  \patchBothAmsMathEnvironmentsForLineno{align}%
  \patchBothAmsMathEnvironmentsForLineno{gather}%
  \patchBothAmsMathEnvironmentsForLineno{multline}%
}

\newcommand{\tablefont}{\fontsize{10}{12}\selectfont\sffamily}

\renewenvironment{abstract}{%
  \vspace{6pt}%
  \begin{center}%
    {\fontsize{14}{16.8}\selectfont\bfseries\sffamily Abstract}%
  \end{center}%
  \vspace{6pt}%
  \fontsize{11}{13.2}\selectfont\sffamily
}{%
  \vspace{12pt}%
}

\newcommand{\papertitle}[1]{%
  \begin{center}
    {\fontsize{16}{19.2}\selectfont\bfseries\sffamily #1\par}
  \end{center}%
  \vspace{12pt}%
}

\newcommand{\paperauthors}[1]{%
  \begin{center}
    {\fontsize{12}{14.4}\selectfont\sffamily #1\par}
  \end{center}%
  \vspace{6pt}%
}

\newcommand{\paperaffiliations}[1]{%
  \begin{center}
    {\fontsize{10}{12}\selectfont\sffamily #1\par}
  \end{center}%
  \vspace{6pt}%
}

\newcommand{\paperemail}[1]{%
  \begin{center}
    {\fontsize{10}{12}\selectfont\sffamily Email for correspondence (presenting author): #1\par}
  \end{center}%
  \vspace{6pt}%
}

\begin{document}
% ============================================================

% Line numbers required for original ATRF conference submission, but arXiv
% rejects submissions containing margin line numbers (submit/7827803,
% "on hold" for line numbers) — disabled for the arXiv copy.
%\linenumbers

% Apply the 3-line conference header on page 1
\thispagestyle{firstpage}

% ------------------------------------------------------------
% TITLE BLOCK
% ------------------------------------------------------------
%\papertitle{A Near-Miss Identification and Prediction Approach Using Connected Vehicles Data}

\papertitle{Proactive Road Safety Intervention in Australia: Predicting Risky Driving Hotspots from Connected Vehicle Data}

\paperauthors{%
  Adriana-Simona Mihăiţă\textsuperscript{1},
  Clarence Cheung\textsuperscript{2},
  Artur Grigorev\textsuperscript{1},
  Tuo Mao\textsuperscript{1},
  David Lillo-Trynes\textsuperscript{3}%
}

\paperaffiliations{%
  \textsuperscript{1}Data Science Institute, University of Technology Sydney (UTS), Australia\\
  \textsuperscript{2}University of Technology Sydney (UTS), Australia\\
  \textsuperscript{3}COMPASS IOT PTY LTD, Australia%
}

\paperemail{adriana-simona.mihaita@uts.edu.au}

% ------------------------------------------------------------
% ABSTRACT (max 300 words)
% ------------------------------------------------------------
\begin{abstract}
Road safety monitoring has historically been reactive, relying on crash-record analysis after fatalities and injuries have already occurred. Proactive identification of high-risk locations and dangerous driving behaviour before incidents occur is a critical but underexplored challenge. This paper addresses this gap using connected vehicle telemetry data from Greater Sydney, Australia, to detect and forecast near-miss risky driving events at the Local Government Area (LGA) level. Risky driving is quantified through g-force thresholds (hard braking $>0.6g$, harsh cornering $>0.47g$, harsh acceleration $>0.5g$), and spatio-temporal heatmaps are constructed to identify high-risk zones. Eight predictive models are benchmarked across three families: ensemble learning (Random Forests, XGBoost, LightGBM), deep learning (LSTM, N-BEATS), and classical time-series methods (ARIMA, Exponential Smoothing, Prophet). ARIMA achieves the lowest mean absolute error (MAE: 162.21), performing comparably to LSTM (MAE: 163.92) and outperforming all ensemble methods, with N-BEATS reaching an MAE of 180.75. These results demonstrate that parsimonious time-series models are competitive with deep learning approaches when training data volume is limited. The study highlights the potential of IoT-based connected vehicle data to support proactive road safety interventions, with Sydney's inner and western LGAs (CBD, Parramatta, Bankstown) identified as persistent high-risk zones warranting targeted policy action.
\end{abstract}

% ============================================================
\section{Introduction}\label{I_Introduction}
% ============================================================

\subsection{Background and motivation}\label{I_A_Background_and_motivation}

Road trauma remains a critical public health issue in Australia. In the 2018--2019 period alone, 39,755 individuals were hospitalised, of whom 10,282 sustained life-threatening injuries \citep{BITRE_2019_safety}. By 2025, Australian road deaths had reached 1,337 — the highest in 12 years — placing the national target of zero fatalities by 2030 under significant pressure. Prior research has identified a range of contributing factors: \citet{cheng_2019_exploring} found associations between crash risk and driver age, experience, road position, time of day, and blood alcohol concentration (BAC). However, the majority of safety interventions remain reactive, triggered only after crash data has been collected. This paper takes a proactive approach, exploiting real-time connected vehicle telemetry to forecast near-miss events before they escalate into crashes.

\subsection{Relevant literature}\label{related_works}

\textbf{Driving behaviour:} Human error is the predominant contributing factor in road incidents \citep{cheng_2019_exploring}. \citet{gitelman_2018_exploring} used in-vehicle data recorders (IVDR) fitted to vehicles driven by novice drivers in Israel, matching recorded braking, acceleration, and speeding events against road infrastructure characteristics and crash data via negative binomial regression. While IVDR is cost-effective and easy to deploy, its event-based capture is coarse and may miss incidents not associated with predefined thresholds.

Naturalistic driving studies (NDS) offer continuous, high-fidelity behavioural recording. \citet{kong_2021_patterns} applied the Apriori algorithm to a US NDS dataset paired with roadway inventory data, finding that near-crash events are strongly associated with roads lacking access control or shoulders, and with speed limits between 30 and 60 mph.

More recently, \citet{li2024connected} applied connected vehicle GPS data (over 2.9~billion GPS points from Wejo) to detect and spatially cluster near-crash events across urban road segments in San Antonio, USA, finding that near-crash risk peaks during weekday commute hours and concentrates in high-density areas — patterns consistent with the Sydney findings we report in this paper. Unlike their Time-To-Collision (TTC) surrogate metric, our work applies g-force thresholds to identify near-miss events, enabling detection using in-vehicle accelerometer data alone without requiring vehicle-to-vehicle proximity information.

The validity of g-force events as near-miss surrogates in the Sydney context is directly substantiated by our recent published work \citet{IEEEITSC25_Artur}, who applied Getis-Ord $G_i^*$ statistics and Bivariate Local Moran's I to a matched dataset of connected vehicle high-g events and official NSW crash records. That study established a statistically significant spatial association between near-miss clusters and crash blackspots — confirming that the g-force proxy captures genuine crash-precursor risk rather than merely reflecting traffic volume, and providing direct empirical grounding for the near-miss detection methodology used in the present paper.

\textbf{Machine learning and deep learning for driving behaviour modelling:}
\citet{wang_2017_driving} applied a semi-supervised support vector machine with $k$-means clustering to classify driving styles (normal, aggressive, jerk, emotional) from continuously recorded vehicle trajectories, demonstrating that ML methods can effectively differentiate behavioural profiles from kinematic data. A systematic review by \citet{elamraniabouelassad_2020_the} confirmed that neural networks, SVMs, decision trees, and ensemble methods are the most widely used approaches for driving behaviour analysis, while noting that results often lack generalisability across datasets. An updated review by \citet{shirole2025review} further documents the growing dominance of gradient-boosting and deep learning architectures in telematics-based driver scoring applications. Addressing the interpretability gap, \citet{masello2023contextual} deployed XGBoost with SHAP explainability to predict near-miss and speeding events from contextual driving data — a methodological approach directly comparable to the ensemble models evaluated in this paper. Beyond individual behaviour, deep learning has been applied to network-level safety problems: \citet{Khaled2022_Vision_transformers} used Vision Transformers to forecast traffic accident risk from contextual scene data, while \citet{Grigorev_2022_NLP_incidents} integrated NLP for feature extraction from incident reports, and prior work by the authors applied deep learning to time-series congestion prediction \citep{Mihaita_2020_Graph_modelling,Mihaita_Arxiv2020}. To the best of our knowledge, the present work is among the first to apply predictive modelling to near-miss \textit{counts} aggregated by LGA, enabling area-level proactive safety intervention rather than post-hoc crash analysis.

\textbf{Innovative contributions of this work compared to the state of art}

The key contributions of this paper are fourfold. First, we present one of the first applications of connected vehicle (IoT) telemetry for near-miss prediction at the Local Government Area (LGA) level in Australia, moving beyond traditional crash-based retrospective analysis. Second, we define a rigorous, data-driven framework for classifying risky driving behaviour using g-force thresholds grounded in an established benchmark study \citep{Ehsani2017}, and apply it to a large-scale real-world dataset spanning over 700,000 vehicles across New South Wales. Third, we conduct a systematic benchmark of eight predictive models spanning three families — ensemble learning, deep learning, and classical time-series — providing practical guidance on model selection for transport safety practitioners working with limited telemetric data. Fourth, we produce actionable, place-based risk maps identifying the Sydney LGAs most prone to dangerous driving, directly supporting targeted road safety interventions by transport agencies and local councils. This work is part of a broader research programme on proactive road safety in Australia: in a related study \citep{Lee2023_TLC_TRB}, we analysed risky driver behaviour near train level crossings using connected vehicle and GIS data, finding that harsh braking at these locations is associated with increased crash risk and informing targeted infrastructure interventions; and in \citet{IEEEITSC25_Artur}, we established a statistically significant spatial association between near-miss clusters and official crash blackspots in Sydney using Getis-Ord $G_i^*$ and Bivariate Local Moran's I analysis, directly validating the g-force proxy used in the present paper.

\subsection{Paper organisation}\label{I_B_Contributions}

This paper is organised as follows: \cref{III_Case_study} introduces the dataset and data mining analysis; \cref{Methodology} presents the modelling framework; \cref{IV_Results} discusses results; and \cref{V_Conclusion} provides conclusions and future directions.

% ============================================================
\section{Case study}\label{III_Case_study}
% ============================================================

The dataset used for this analysis was provided by a transportation data provider Compass IoT \citep{CompassIoT} operating in Australia. This organisation collects telemetric information from connected vehicles, including real-time movement records at second-level resolution, origin-destination traffic volumes, travel times, and speed metrics.
\textbf{For the current study, two event types were utilised:
(a) \textit{safe-point data}, which identifies events based on accelerometer readings (including harsh vehicle movements, violent manoeuvres, near-misses, collisions, or high g-force incidents), and
(b) \textit{brake-point data}, which records locations of collective acceleration/ deceleration patterns to identify infrastructure characteristics and congestion points.}

Data collection occurs via in-vehicle Data Acquisition Systems (DAS) with cloud-based transmission. The dataset for this study comprises over 700,000 vehicles from 64 manufacturers (April 2020 until December 2021), with top 1\% g-force events removed for statistical consistency. This processing eliminates extreme values that may represent measurement errors or faulty devices. As the data is collected directly by vehicle systems without external hardware requirements, it does not include video or audio verification of incidents. The dataset represents recorded driving behaviour patterns rather than verified crash events.

\subsection{Definitions}

\textbf{Definition 1:} A g-force measures acceleration relative to gravitational acceleration (1g = 9.806 m/s²).

\begin{figure}[h!]
    \caption{The x/y/z axis definition.}
    \label{XYZ}
    \centering
    \includegraphics[scale=0.5]{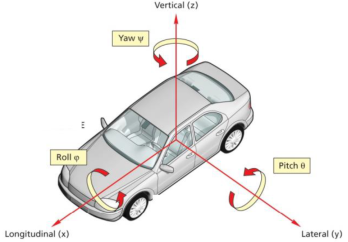}
\end{figure}

\textbf{Definition 2:} The X/Y/Z-axis convention is specified in ISO 8855:1991, in which the x-axis points towards the front of the vehicle, the y-axis towards the left, and the z-axis upwards (right-hand system), with the origin at the most forward point on the centre-line of the vehicle for dynamic data measurements (as shown in \cref{XYZ}).

\textbf{Definition 3:} The severity of the g-force is represented as harsh acceleration and harsh braking. Deceleration or braking from 60 km/h to 0 in 3.0 seconds at 0.6 g is considered harsh braking. Accelerating from 0 to 60 km/h over 0.5 g's is considered a hard acceleration. Swerving or cornering is considered harsh when it is over 0.47 g's. The g-force thresholds are shown in Table \ref{tab:gforce}, following the benchmark study of \citet{Ehsani2017} and consistent with the threshold validation in connected-vehicle telematics by \citet{hossain2026telematics}. The spatial correspondence between events identified by these thresholds and official crash blackspots in Sydney has been confirmed by \citet{IEEEITSC25_Artur}; together, these studies ground the near-miss classification used in this work.

\begin{table}[h!]
\caption{G-force classification thresholds for near-miss event detection adopted in this study, following \citet{Ehsani2017} and validated against connected-vehicle telematics by \citet{hossain2026telematics}.}
\label{tab:gforce}
\centering
\tablefont
\begin{tabular}{|l|c|}
\hline
\textbf{Driving Manoeuvre} & \textbf{Classification threshold} \\
\hline
\hline
Hard braking & $> 0.6$ G \\
\hline
Harsh cornering & $> 0.47$ G \\
\hline
Harsh acceleration & $> 0.5$ G \\
\hline
\end{tabular}
\end{table}

% \begin{table}[h!]
% \caption{Target g-force reference ranges for various driving manoeuvres \citep{Ehsani2017}. Note: the specific classification thresholds adopted in this study are defined in Definition~3 (hard braking $>0.6g$, harsh cornering $>0.47g$, harsh acceleration $>0.5g$).}
% \label{tab:gforce}
% \centering
% \tablefont
% \begin{tabular}{|l|c|}
% \hline
% \textbf{Driving Manoeuvre} & \textbf{Target g-force} \\
% \hline
% \hline
% Cornering & \\
% \hline
% \quad Mild Left & 0.2 - 0.3 G's \\
% \hline
% \quad Moderate Left & 0.3 - 0.4 G's \\
% \hline
% \quad Hard Left & 0.5 - 0.6 G's \\
% \hline
% \quad Mild Right & 0.2 - 0.3 G's \\
% \hline
% \quad Moderate Right & 0.3 - 0.4 G's \\
% \hline
% \quad Hard Right & 0.5 - 0.6 G's \\
% \hline
% Braking & \\
% \hline
% \quad Mild & 0.4 - 0.5 G's \\
% \hline
% \quad Moderate & 0.5 - 0.6 G's \\
% \hline
% \quad Hard & 0.6 - 0.7 G's \\
% \hline
% Acceleration from stationary position & \\
% \hline
% \quad Mild & 0.2 G's \\
% \hline
% \quad Hard & 0.3 - 0.4 G's \\
% \hline
% Turns & \\
% \hline
% \quad Mild Left & 0.2 - 0.3 G's \\
% \hline
% \quad Moderate Left & 0.4 - 0.5 G's \\
% \hline
% \quad Hard Left & 0.6 - 0.7 G's \\
% \hline
% \quad Mild Right & 0.2 - 0.3 G's \\
% \hline
% \quad Moderate Right & 0.4 - 0.5 G's \\
% \hline
% \quad Hard Right & 0.6 - 0.7 G's \\
% \hline
% \end{tabular}
% \end{table}

\subsection{Data mining and analytics}\label{Data_Mining_and_Analytics}

\begin{figure}[h!]
    \caption{Driving incident classifications.}
    \label{incident_classifications}
    \centering
    \includegraphics[scale=0.5]{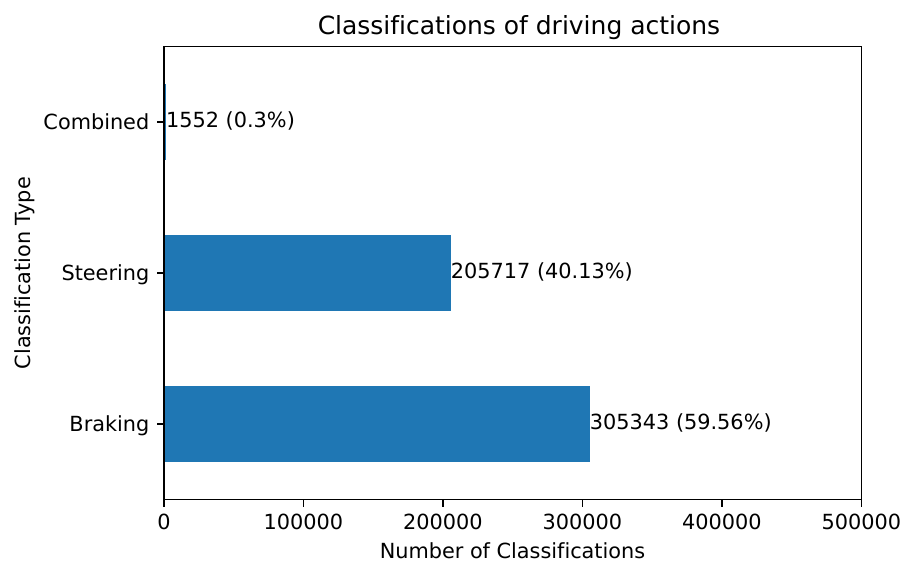}
\end{figure}

The first step is to understand what type of driving actions have been recorded the most inside the data set; as shown in \cref{incident_classifications} most of the actions are either ``Steering''(40.13\%) or ``Braking''(59.56\%), with very few as combined events (0.3\%). This classification has been done based on incoming Compass IoT data sets that record the vehicle manoeuvres while the risky driving behaviour has occurred (e.g. braking while also steering the vehicle wheel to the left is seen as a combined event, whereas simple steering without braking is a turning movement of the vehicle on the road).

% --- ORIGINAL (two separate figures — merged below) ---
%We further analyze on what type of road the risky driving behaviour has occurred (see \cref{road_classif}) and we identify that trunk roads (connecting cities) are the ones that carry most of the risky driving behaviour due to increased traffic volumes (30.3\%), followed by primary roads such as highways, freeways or tollways (25.2\%) and secondary roads such as neighborhood roads (16.6\%).
%
%\begin{figure}[h!]
%    \caption{Road classification distribution of incidents.}
%    \label{road_classif_old}
%    \centering
%    \includegraphics[scale=0.5]{Road_Class_Pie.pdf}
%\end{figure}
%
%When it comes to the number of lanes that incidents occurred on, our investigation revealed that most of them affect only one lane on 51.2\%, followed by two lanes (22.8\%) or three lanes (19.4\%) - (see \cref{nr_lanes}).
%
%\begin{figure}[h!]
%    \caption{Road classification distribution.}
%    \label{nr_lanes_old}
%    \centering
%    \includegraphics[scale=0.5]{Lane_Count_Pie.pdf}
%\end{figure}
% --- END ORIGINAL ---

\begin{figure}[h!]
    \caption{Road infrastructure characteristics of risky driving incidents.
    (a)~Distribution by road classification: trunk roads account for the largest share (30.3\%), followed by primary (25.2\%) and secondary roads (16.6\%).
    (b)~Distribution by number of lanes: single-lane roads account for 51.2\% of incidents, followed by two-lane (22.8\%) and three-lane (19.4\%) roads.}
    \label{fig:road_lanes}
    \centering
    \begin{subfigure}[t]{0.5\linewidth}
        \centering
        \includegraphics[width=\linewidth]{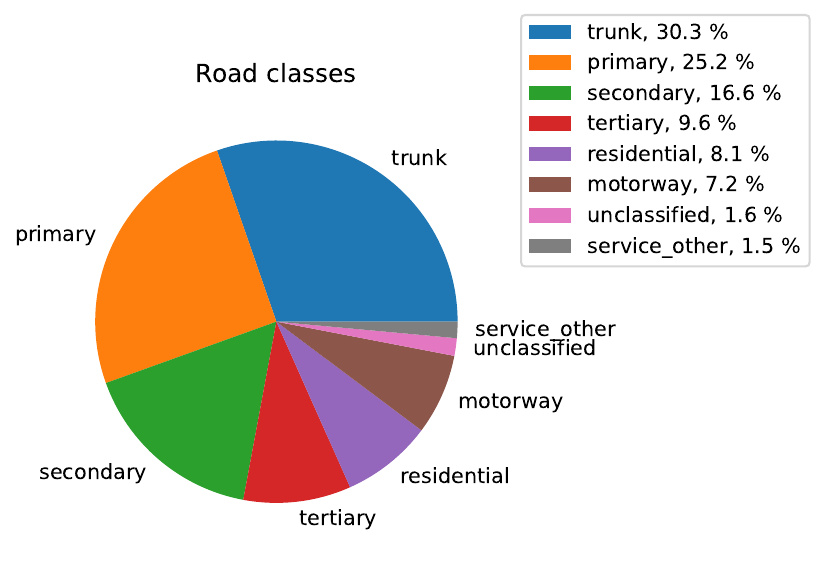}
        \subcaption{Road classification distribution.}
        \label{road_classif}
    \end{subfigure}\hfill
    \begin{subfigure}[t]{0.42\linewidth}
        \centering
        \includegraphics[width=\linewidth]{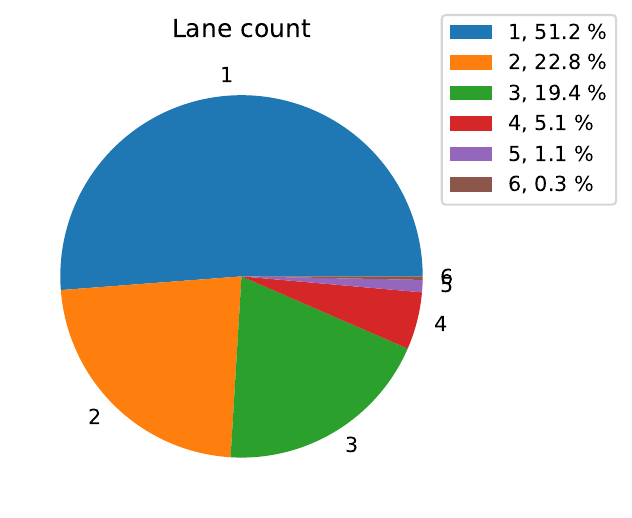}
        \subcaption{Lane count distribution.}
        \label{nr_lanes}
    \end{subfigure}
\end{figure}

We further analyse the road type and lane configuration of incidents in \cref{fig:road_lanes}. \Cref{road_classif} shows that trunk roads (connecting cities) carry the most risky driving behaviour due to higher traffic volumes (30.3\%), followed by primary roads such as highways, freeways or tollways (25.2\%), and secondary roads such as neighbourhood roads (16.6\%). \Cref{nr_lanes} reveals that most incidents affect only a single lane (51.2\%), followed by two-lane (22.8\%) and three-lane (19.4\%) roads.

\begin{figure}[h!]
    \caption{Road classification distribution by lane types.}
    \label{Road_classification_distribution}
    \centering
    \includegraphics[width=\linewidth]{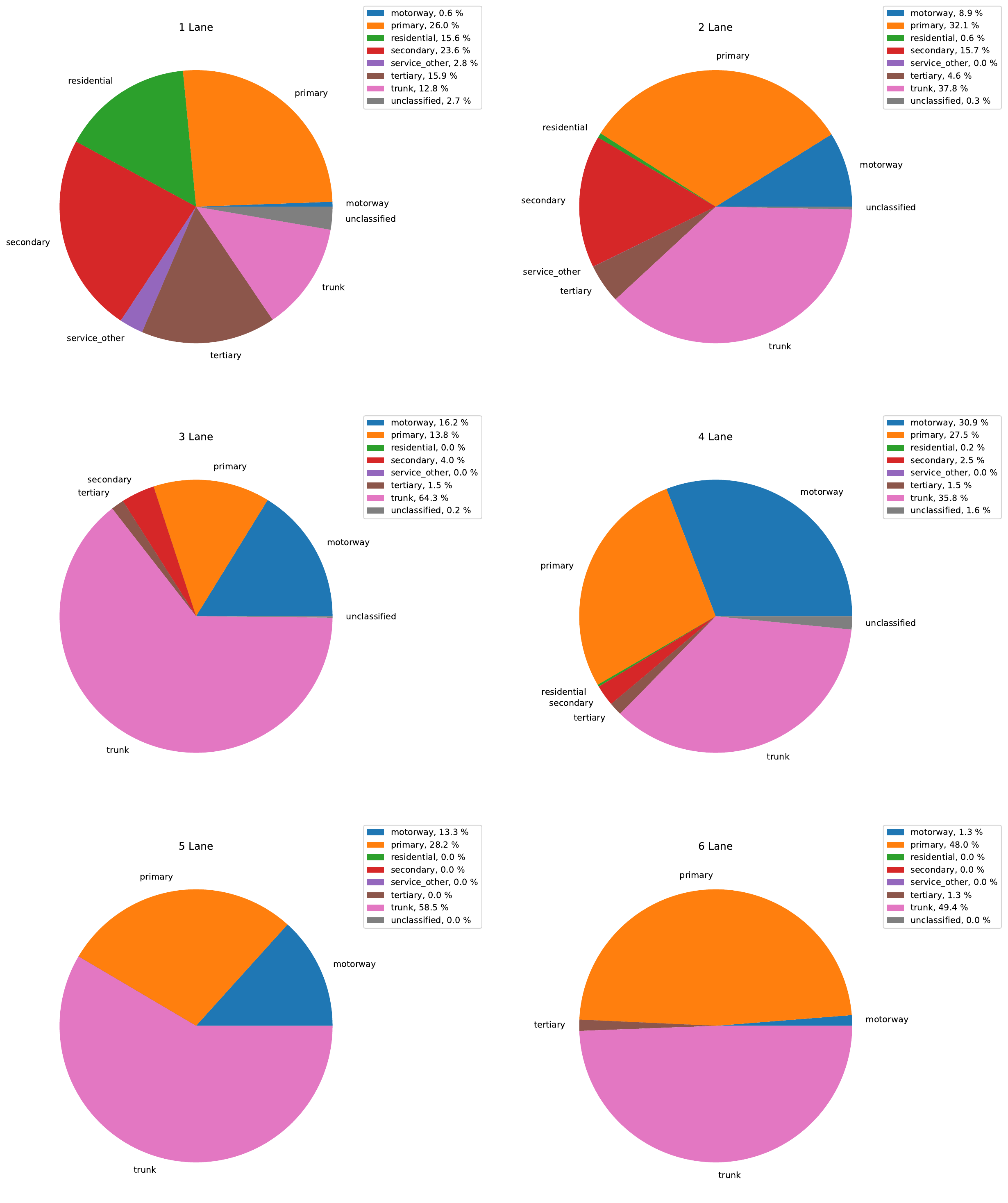}
\end{figure}

\Cref{Road_classification_distribution} further breaks down incidents by road class and lane count, revealing that 1-lane roads are predominantly primary, secondary, residential, and tertiary roads — the road types most commonly associated with risky driving incidents in this dataset.

\begin{figure}[h!]
    \caption{Probability distribution function of the driving speed.}
    \label{speed_distribution}
    \centering
    \includegraphics[scale=0.5]{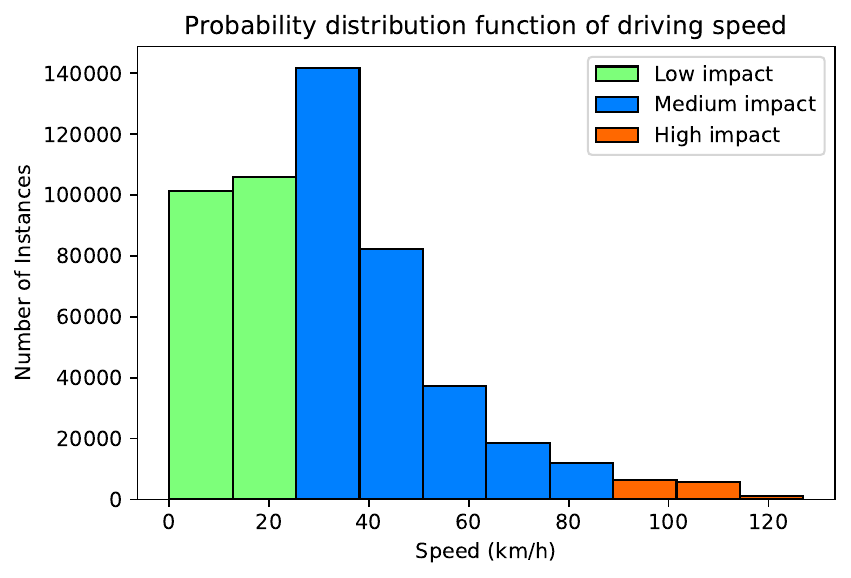}
\end{figure}

Next, we further analyse the probability distribution of the recorded speed during risky driving behaviour and categorise it into three classes: low, medium, and high impact (see \cref{speed_distribution}). We observe that most of the risky driving behaviour happens at low speeds (less than 20km/h; green) while the most frequent events happen at medium speeds between 25-40km/h (blue). There are however few risky events that arrive at speeds higher than 100km/h but these tend to occur less often.

\begin{figure}[h!]
    \caption{Weekly profiling of incidents.}
    \label{weekly_profiling}
    \centering
    \includegraphics[width=\linewidth,height=0.9\textheight,keepaspectratio]{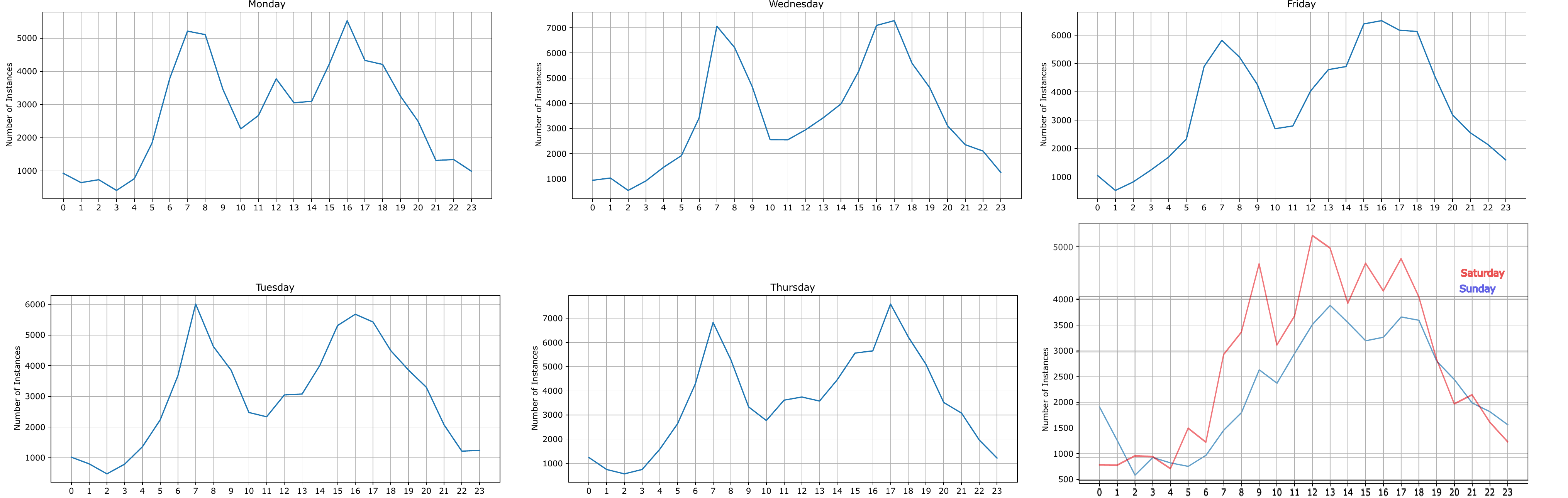}
\end{figure}

The daily trend analysis of incidents (see \cref{weekly_profiling}) reveals that most events arrive during the weekdays' morning and afternoon peak hours (7-9 AM and 3-6 PM), especially on Wednesdays and Fridays when almost 7,000 risky driving behaviour events were recorded at the peak hours. The weekend usually has less risky driving events than the weekdays and most of these occur during the afternoon time, which corresponds to regular outside work hours and weekend leisure activities.

Another important part of our work is understanding what are the local government areas (LGAs) that are the most affected by a risky driving behaviour and whether this can be predicted in the future using machine learning approaches.

\begin{figure}[h!]
    \caption{Mapping of driving incidents.}
    \label{Mapping}
    \centering
    \includegraphics[width=0.6\linewidth]{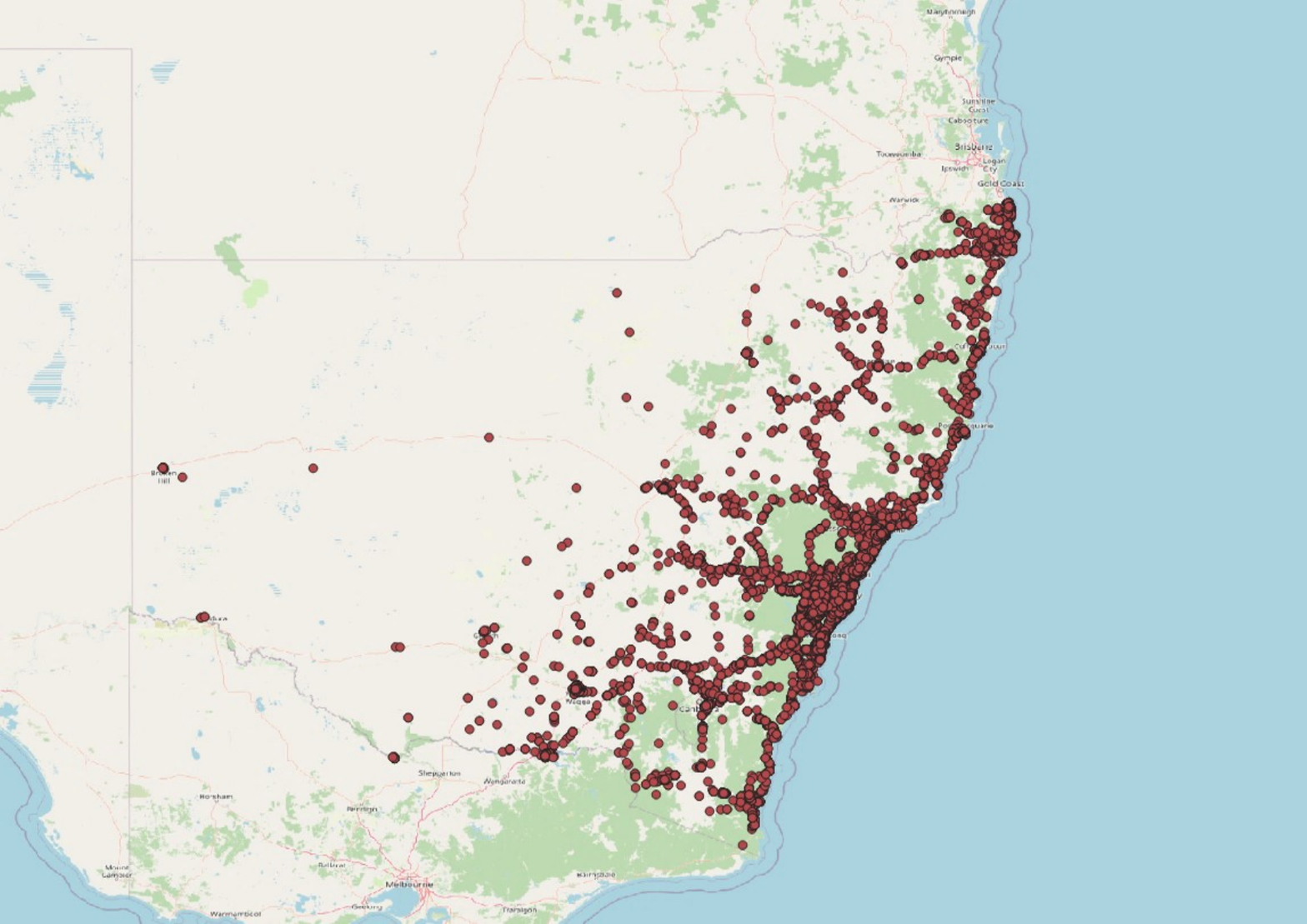}
\end{figure}

\begin{figure}[h!]
    \caption{Heatmap of driving incidents.}
    \label{heatmap}
    \centering
    \includegraphics[width=0.6\linewidth]{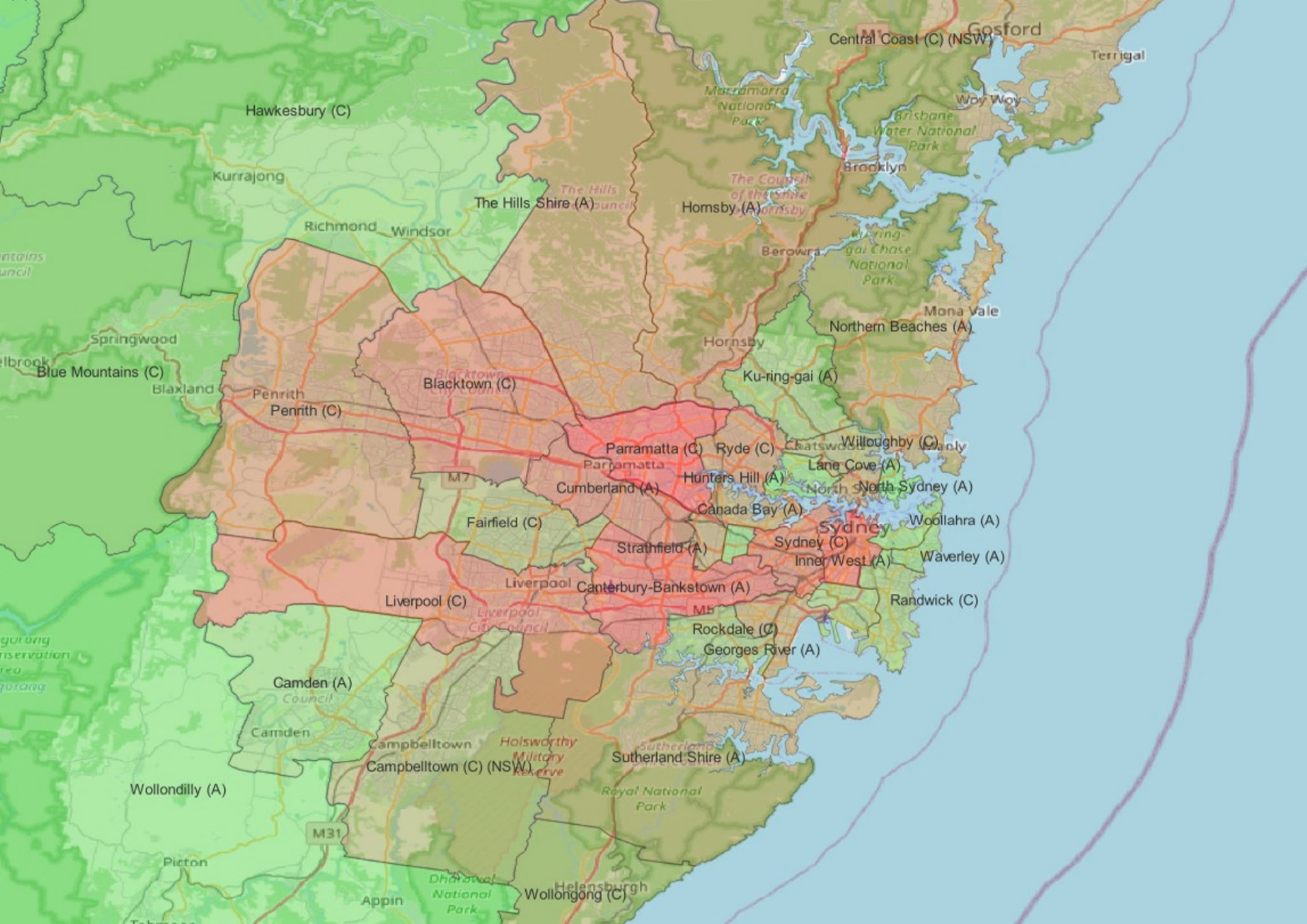}
\end{figure}

\cref{Mapping} shows the geo-spatial mapping of incidents mostly inside the NSW region (with some of them in the Queensland region). Most of the risky driving events occur in areas close to cities while only a few in the countryside which can be explained by a lower population density and less complex infrastructure. Based on this mapping and the grouping of incidents by LGAs, we plot the heatmap of incidents in \cref{heatmap}, which indicates that red areas have high incident counts while green areas have lower occurrences.

\begin{figure}[h!]
    \caption{Ranking of urban LGAs across the city of Sydney.}
    \label{LGA_rank}
    \centering
    \includegraphics[scale=0.5]{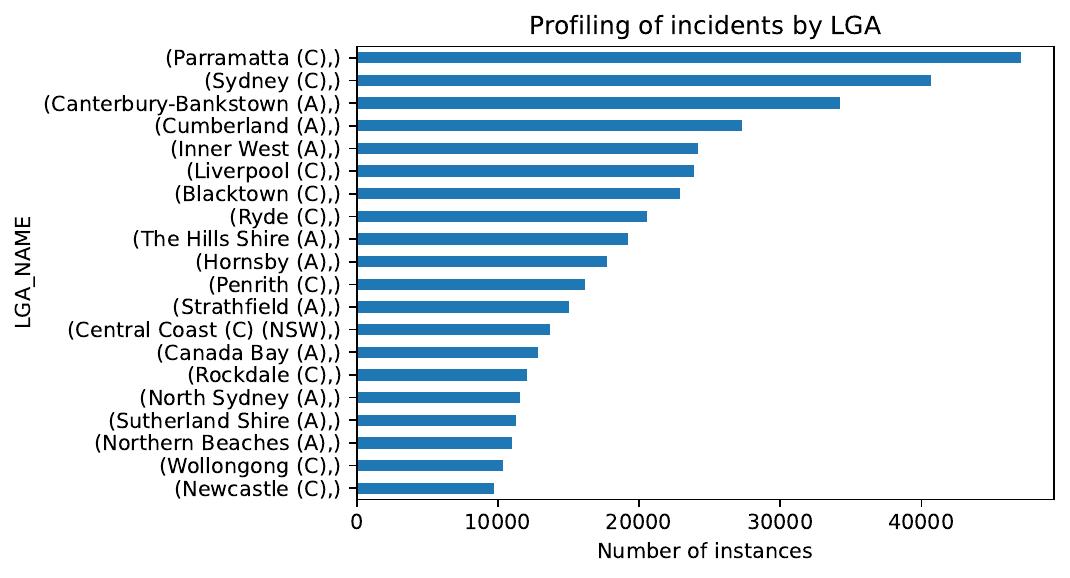}
\end{figure}

\cref{LGA_rank} shows the ranking of LGAs based on the total number of risky driving behaviour incidents and the most dangerous areas seem to be located in the inner and western part of the city of Sydney (CBD, Parramatta and Bankstown), in areas with a high population density and busy roads (high traffic volumes).

It is acknowledged that LGA-level incident counts reflect both the prevalence of risky driving behaviour and underlying traffic exposure. However, the near-miss spatial signal is demonstrably not reducible to a traffic-volume proxy. \citet{IEEEITSC25_Artur} applied the same Compass IoT dataset at a fine 400m grid resolution and found that the Global Bivariate Moran's I between near-miss counts and official crash counts was 0.1654 ($p < 0.001$) — statistically significant but moderate, confirming that the two indicators are related yet distinct. Critically, 2,681 grid cells across Sydney exhibited a Low-Crash, High-Near-Miss (LH) pattern: locations with statistically elevated near-miss frequency but low historical crash counts. These are areas of latent risk that would be invisible to crash-only analysis and would be lost under naive exposure normalisation. The existence of these LH areas at scale demonstrates that the near-miss metric captures genuine behavioural risk over and above what traffic volume alone can explain. Normalisation by vehicle-kilometres travelled or active vehicle counts per LGA remains a planned extension of this work as fleet-level trip data becomes available from Compass IoT.

% --- ORIGINAL (two separate figures — merged below) ---
%\cref{pdf_LGAs} shows the probability distribution of the daily number of incidents per day, reaching on average 1,000 incidents with a maximum number of 3,000. We also observe that the pdf does not have a long tail which indicates that the prediction problem is highly achievable in terms of accuracy.
%
%\begin{figure}[h!]
%    \caption{Probability Distribution Function of daily incidents.}
%    \label{pdf_LGAs_old}
%    \centering
%    \includegraphics[scale=0.5]{Incidents_Per_Day.pdf}
%\end{figure}
%
%Furthermore, \cref{ecdf} shows the empirical cumulative distribution function of incidents in each LGA. This graph shows that many LGAs have zero incidents recorded and roughly 20\% of all LGAs have had more than 10,000 incidents which are most likely connected to the top areas identified previously in \cref{LGA_rank}.
%
%\begin{figure}[h!]
%    \caption{ECDF of number of incidents in each LGA.}
%    \label{ecdf_old}
%    \centering
%    \includegraphics[scale=0.5]{ECDF.pdf}
%\end{figure}
% --- END ORIGINAL ---

\begin{figure}[h!]
    \caption{Distribution of daily near-miss incidents across NSW LGAs.
    (a)~Probability Distribution Function (PDF) of daily incident counts, averaging approximately 1,000 per day with a maximum of around 2,000.
    (b)~Empirical Cumulative Distribution Function (ECDF) of total incident counts per LGA, showing that roughly 20\% of LGAs account for more than 10,000 incidents.}
    \label{fig:pdf_ecdf}
    \centering
    \begin{subfigure}[t]{0.48\linewidth}
        \centering
        \includegraphics[width=\linewidth]{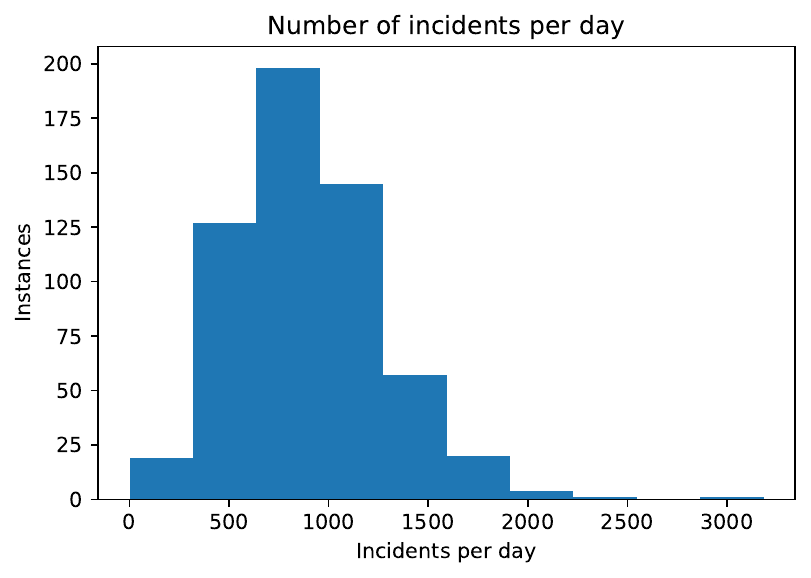}
        \subcaption{PDF of daily incidents.}
        \label{pdf_LGAs}
    \end{subfigure}\hfill
    \begin{subfigure}[t]{0.48\linewidth}
        \centering
        \includegraphics[width=\linewidth]{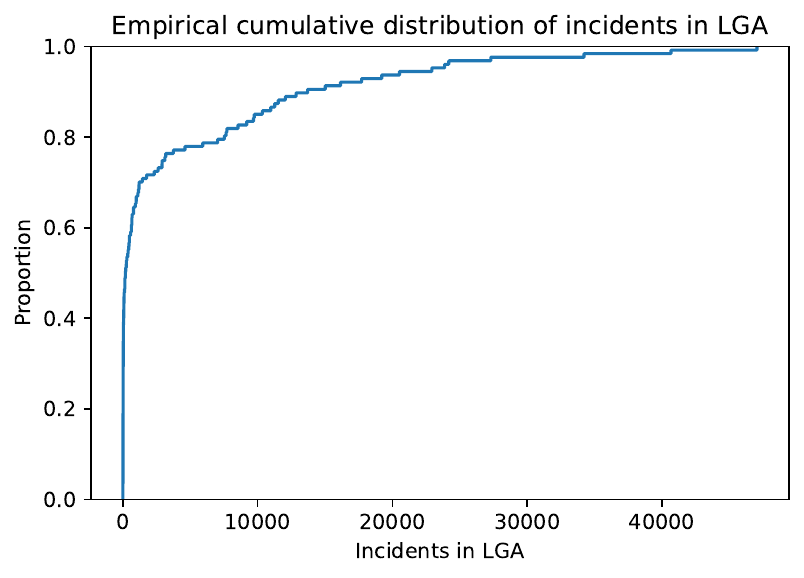}
        \subcaption{ECDF of incidents per LGA.}
        \label{ecdf}
    \end{subfigure}
\end{figure}

\Cref{pdf_LGAs} shows the probability distribution of the daily number of incidents, averaging approximately 1,000 per day with a maximum of around 2,000. The distribution does not exhibit a long tail, indicating that the range of values to forecast is well-constrained and the prediction problem is tractable. \Cref{ecdf} shows the empirical cumulative distribution function (ECDF) of total incidents per LGA: many LGAs have zero incidents recorded, and roughly 20\% of all LGAs have experienced more than 10,000 incidents — consistent with the high-risk areas identified in \cref{LGA_rank}.

% ============================================================
\section{Methodology}\label{Methodology}
% ============================================================

An important objective of our work is to test whether the past historical logs of risky driving behaviour can help predict the number of incidents likely to occur in each Local Government Area on the following day, and thereby identify which LGAs will be the riskiest in the near future. Critically, a separate model is trained independently for each LGA, treating each LGA's daily incident count as its own univariate time series. This per-LGA approach enables area-level forecasts that can directly inform targeted resource deployment and preventive measures by transport authorities.

Several machine learning (ML) and deep learning (DL) models have been deployed for solving this challenge as detailed below.

All models operate on the same univariate input: the daily incident count time series per LGA. No external features (road geometry, speed limits, weather, or infrastructure type) were incorporated at the forecasting stage; the descriptive road-attribute analysis in \cref{Data_Mining_and_Analytics} characterises the dataset but does not enter the predictive models. For the regression-based ensemble models (RF, XGBoost, LightGBM), the univariate series is converted to a supervised learning problem via a lagged-feature representation: the past 30 daily observations per LGA are used as input features to predict the next day's incident count. The time-series models (ARIMA, ES, Prophet) and neural network models (LSTM, N-BEATS) consume the same historical window through their respective sequential architectures. The comparison across all eight models is therefore fair and consistent.

\textbf{Evaluation protocol.} All models are evaluated using a rolling origin (expanding-window) procedure implemented via the Darts library \texttt{historical\_forecasts} function \citep{herzen2022darts}. The held-out evaluation window comprises the final 35 days of each LGA's time series; the preceding observations constitute the initial training set. Evaluation proceeds sequentially across the 35-day window, generating one-step-ahead forecasts at each daily time step — this constitutes a standard rolling origin evaluation with 35 forecast origins, rather than a single static held-out prediction. For all statistical and ensemble models (ARIMA, ES, Prophet, RF, XGBoost, LightGBM), the model is retrained at every evaluation step using all observations available up to that point (expanding window), ensuring that no future data is ever used. For the neural network models (LSTM, N-BEATS), the model is trained once on the initial portion and then evaluated with rolling one-step-ahead forecasts across the 35-day window. Zero-incident daily observations are replaced by a count of one prior to evaluation to prevent undefined MAPE values in LGAs with sparse records.

a) \textbf{Random Forests (RF)}, also known as random decision forests, utilise bootstrap-aggregation (bagging), wherein models are trained on randomly chosen data subsets. It then combines the average or majority votes of multiple decision trees to mitigate the impact of noise on a single tree model. Their utilisation is well spread among different types of research problems in the transport and smart mobility areas \citep{Mihaita2019a}, \citep{Mihaita2019}.

b) \textbf{XGBoost} \citep{chen2016xgboost} is a type of extreme gradient boosting trees which utilises an exhaustive exploration of split values by considering all possible splits across all features. Furthermore, it incorporates a regularisation parameter within the objective function. They have been used widely in several predictions tasks for incident duration modelling, for example, and their performance has exceeded so far the baseline ML models (see \citep{Mihaita2019}, \citep{Grigorev_ITSC2022_Duration_prediction_DL_NLP}).

c) \textbf{LightGBM} \citep{NIPS2017_6449f44a} is a highly efficient and scalable implementation of the gradient boosting algorithm, designed to provide faster training speed and lower memory usage when compared to other gradient boosting frameworks. It achieves this by utilising a histogram-based approach for finding the best split points during the tree-building process. LightGBM also supports various advanced features such as handling categorical features, custom loss functions, and parallel training and has been successfully deployed in disruption modelling studies \citep{GRIGOREV2022_TRC}.

d) \textbf{ARIMA} (Autoregressive Integrated Moving Average) is a widely used time series forecasting model \citep{box1994time} which makes predictions for the next value in a series by combining past observations in a linear manner. We used a default ARIMA implementation with hyperparameters (p = 2, d = 1, q = 0). In this context, p represents the autoregressive parameter, d signifies the degree of differencing (indicating the number of past value subtractions), and q controls the moving average component.

We determined the best hyperparameter values using an exhaustive line search on the validation set, exploring the domains $p \in \{1,..,5\}$, $d \in \{1,..,5\}$, and $q \in \{0,..,3\}$. It is worth noting that ARIMA can only predict a single future value (see other examples in \citep{Sajjad_2022_transportation_Letters}).

e) \textbf{Exponential Smoothing (ES)} assigns exponentially decreasing weights to past observations \citep{gardner1985exponential}. This study uses the Triple Exponential Smoothing (Holt-Winters) variant, which jointly models level, trend, and seasonality.

f) \textbf{LSTM} (Long Short-Term Memory) model is a type of recurrent neural network (RNN) architecture that is specifically designed to overcome the limitations of traditional RNNs in capturing long-term dependencies in sequential data. LSTM networks utilise memory cells and various gating mechanisms to selectively remember and forget information over long sequences. This allows LSTMs to effectively model and predict sequences with complex temporal dependencies \citep{hochreiter1997long}. The model has been used with success in several traffic flow prediction problems \citep{Mihaita_Arxiv2020, Grigorev_2022_NLP_incidents} and for the training we have used 80 hidden units and a ReLU activation function. In general training the LSTM needs at least 15 epochs to converge and the loss function is MSE (mean squared error).

g) \textbf{Prophet} \citep{taylor2018forecasting} is a decomposable time series model comprising trend, seasonality, and holiday components estimated via a Bayesian framework. Hyperparameters (seasonality settings, changepoint parameters, and NSW public holidays) were optimised via cross-validation and grid search.

% g) \textbf{Prophet} is a forecasting model developed by Facebook's Core Data Science team, specifically designed for time series analysis. It incorporates additive regression models that capture various seasonal patterns, as well as holiday effects. Prophet employs a decomposable time series model, consisting of trend, seasonality, and holiday components, and uses a Bayesian framework to estimate these components. It also provides flexibility in handling missing data and outliers. Prophet has gained popularity due to its ease of use, automatic handling of multiple seasonality, and ability to generate reliable and interpretable forecasts \citep{taylor2018forecasting}. Some of the parameters that we optimised via cross validation and grid search are seasonality settings, changepoint parameters, and handling of holidays (via a specification of holiday days in NSW).

h) \textbf{N-BEATS} (Neural Basis Expansion Analysis for Interpretable Time Series Forecasting) is a deep learning model specifically designed for time series forecasting. It aims to provide interpretable and accurate predictions by decomposing the forecasting task into a series of basis functions. N-BEATS utilises a stack of fully connected neural networks, each learning a specific basis function whose outputs are combined to generate the final forecast. N-BEATS has shown promising results in terms of interpretability and forecasting accuracy across various domains \citep{oreshkin2020nbeats}. The model has been trained using the Stochastic gradient Descent algorithm, and the hyper-parameter optimization has been done via grid search of several configurations of the number of stacks, number of layers, learning rate, batch size, or the regularisation technique.

\subsection{Model evaluation metrics}

Three standard metrics are used to evaluate model performance: RMSE (Root Mean Square Error), MAE (Mean Absolute Error), and MAPE (Mean Absolute Percentage Error).

\(RMSE = \sqrt{\frac{1}{n}\sum_{i=1}^{n}{\Big(y_i - \hat{y}_i\Big)^2}}\)

\(MAE = \frac{1}{n}\sum_{i=1}^{n}\left | y_{i} - \hat{y}_{i} \right |\)

\(MAPE = \frac{1}{n}\sum_{i=1}^{n}\left |\frac{ y_i - \hat{y}_i}{y_i} \right |\)

% ============================================================
\section{Results}\label{IV_Results}
% ============================================================

\begin{table}[h!]
    \caption{Prediction error results.}
    \label{tab_results}
    \centering
    \tablefont
    \begin{tabular}{lccc}
    \toprule
     Method     & MAE       & RMSE          & MAPE\\
    \midrule
     Ensemble Learning Models     &        &           & \\
        RF      & 179.15     & 223.81       & 22.53\\
    XGBoost     & 227.19    & 281.71        & 28.27  \\
    LightGBM    & 182.65     & 231.70       & 22.49  \\
    \midrule
         Time Series Models     &        &           & \\
    \textbf{ARIMA}       & \textbf{162.21}     & \textbf{206.07}       & \textbf{20.41}  \\
    ES          & 168.41     & 214.67       & 21.54 \\
    Prophet     & 236.84     & 270.79       & 27.69\\
    \midrule
         Neural Network Models     &        &           & \\
    LSTM        & 163.92     & 225.16       & 21.08 \\
    N-BEATS     & 180.75     & 229.34       & 22.19 \\
    \bottomrule
    \end{tabular}
\end{table}

The results of all models are presented in \cref{tab_results}, grouped into three classes: ensemble learning (RF, XGBoost, LightGBM), time-series models (ARIMA, ES, Prophet), and neural network models (LSTM, N-BEATS). Each model was trained and evaluated independently for each LGA; the reported MAE, RMSE, and MAPE values are averaged across all LGAs.

\begin{figure}[ht]
    \caption{Rolling origin forecast results over the 35-day evaluation window.
    Top row: best-performing models — (a)~ARIMA (MAE: 162.21) and (b)~LSTM (MAE: 163.92) — closely tracking observed incident counts.
    Bottom row: poorest performers — (c)~XGBoost (MAE: 227.19) and (d)~Prophet (MAE: 236.84) — showing systematic prediction errors.}
    \label{fig:model_forecasts}
    \centering
    \begin{subfigure}[t]{0.48\linewidth}
        \centering
        \includegraphics[width=\linewidth]{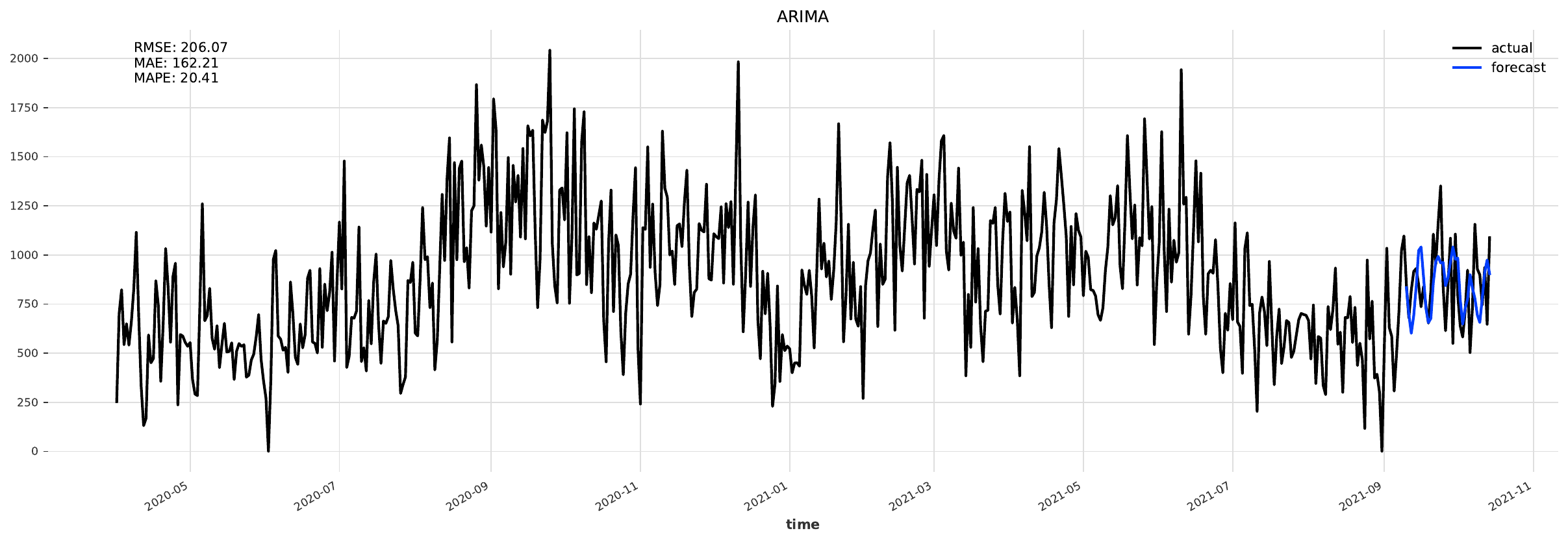}
        \subcaption{ARIMA — best overall.}
        \label{ARIMA_best}
    \end{subfigure}\hfill
    \begin{subfigure}[t]{0.48\linewidth}
        \centering
        \includegraphics[width=\linewidth]{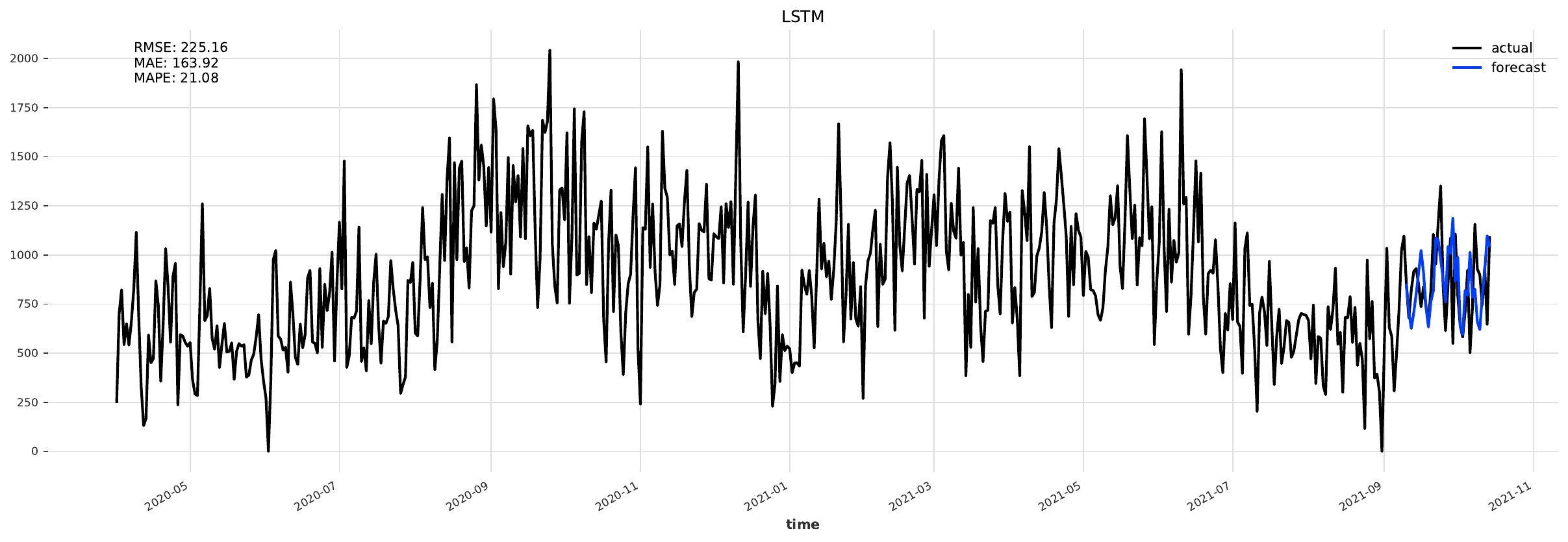}
        \subcaption{LSTM — second best.}
        \label{LSTM_second}
    \end{subfigure}

    \medskip

    \begin{subfigure}[t]{0.48\linewidth}
        \centering
        \includegraphics[width=\linewidth]{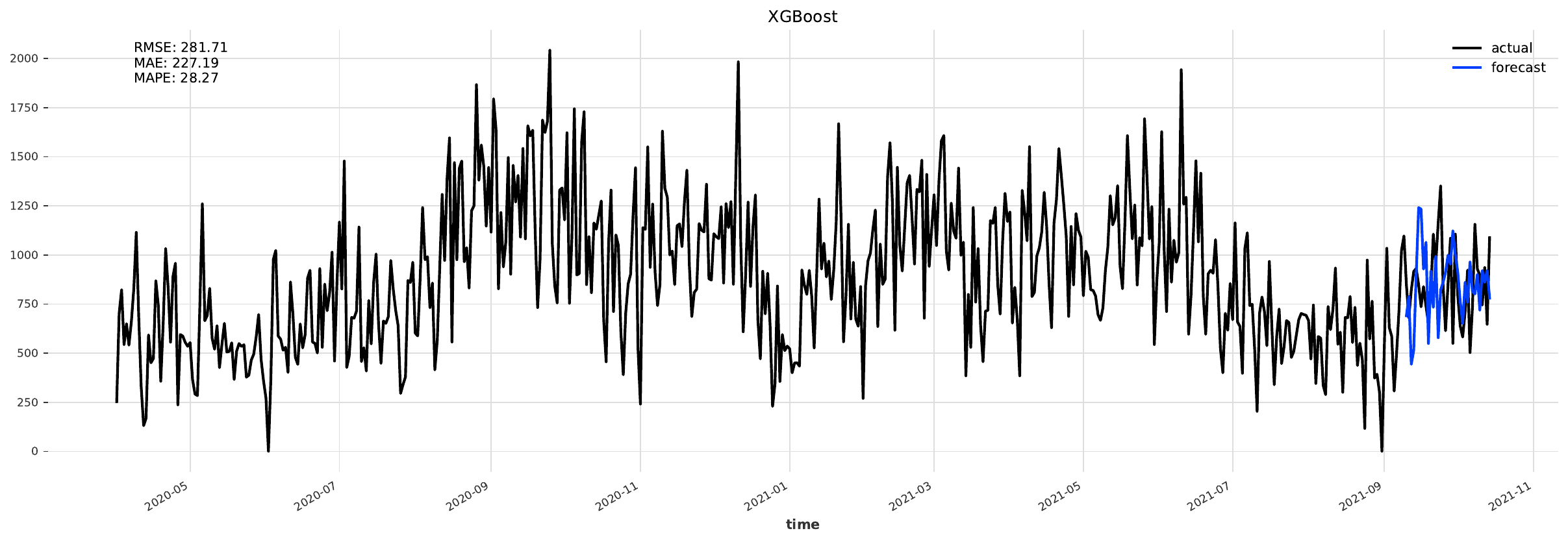}
        \subcaption{XGBoost — worst ensemble model.}
        \label{XGBoost_worst}
    \end{subfigure}\hfill
    \begin{subfigure}[t]{0.48\linewidth}
        \centering
        \includegraphics[width=\linewidth]{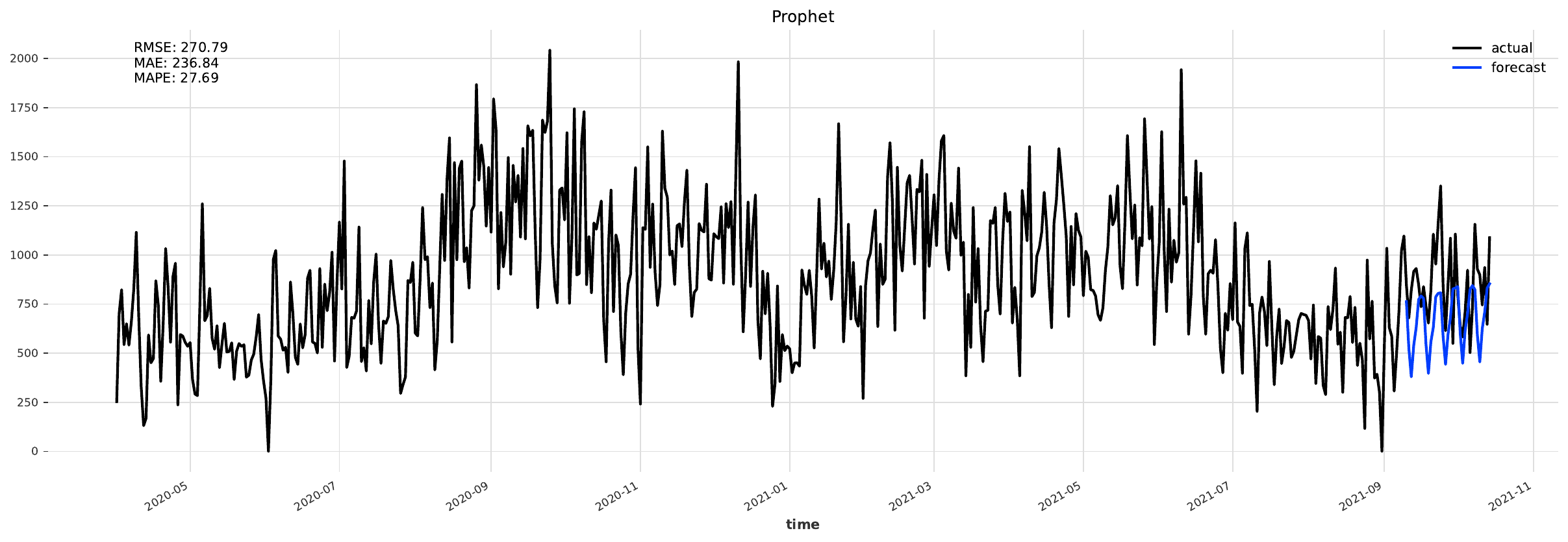}
        \subcaption{Prophet — worst time-series model.}
        \label{Prophet_bad}
    \end{subfigure}
\end{figure}

Findings reveal that ensemble learning models performed among the worst, while traditional time-series models and neural network architectures achieved comparable accuracy. ARIMA achieved the lowest error (MAE: 162.21), with LSTM performing at a very similar level (MAE: 163.92); without a formal statistical comparison such as the Diebold-Mariano test, these two models should be regarded as comparable rather than definitively ranked. Overall, Prophet, XGBoost, and LightGBM were among the weakest models for this dataset (see \cref{XGBoost_worst,Prophet_bad}), while ARIMA, ES, and LSTM were the top three performing models with comparable metrics.

\Cref{fig:model_forecasts} compares rolling origin forecasts across four selected models over the 35-day evaluation window. The top row shows the two best performers — ARIMA \cref{ARIMA_best} and LSTM \cref{LSTM_second} — both of which track the observed series closely (MAE: 162.21 and 163.92 respectively). The bottom row shows the two poorest performers — XGBoost \cref{XGBoost_worst} and Prophet \cref{Prophet_bad} — where systematic over- or under-prediction is clearly visible throughout the evaluation window (MAE: 227.19 and 236.84 respectively).

The observed strong performance of ARIMA, a traditional time-series model, relative to more modern ML/DL counterparts can be attributed to several factors. First, the prediction task possesses relatively simple temporal dependencies and linear patterns — characteristics that align well with ARIMA's design for univariate time-series forecasting. All models in this study, including the ensemble methods, receive only the univariate incident count series as input (via 30-day lagged features for RF, XGBoost, and LightGBM); the absence of additional contextual features means that the ensemble models' capacity for non-linear interaction modelling cannot be fully exploited, which likely contributes to their underperformance relative to the more parsimonious ARIMA specification. While ARIMA and LSTM achieve very similar errors (MAE: 162.21 vs.\ 163.92), it must be noted that their predictive capacity was evaluated on short-term (single-step ahead) forecasting, which may show different relative performance on longer evaluation horizons.

The results come as a surprise that more complex deep learning models are not well suited for the current research problem in question, as depicted by the data set, and that the traditional models work the best. However, this can be related to several limitations of our studies as detailed in the Conclusions section.

% ============================================================
\section{Conclusions}\label{V_Conclusion}
% ============================================================

This paper presented a data-driven framework for the identification and short-term forecasting of near-miss risky driving events in New South Wales, Australia, using connected vehicle (IoT) telemetry data from over 700,000 vehicles. By defining risky driving through g-force thresholds and aggregating events at the Local Government Area (LGA) level, we demonstrated that spatial patterns of dangerous behaviour are highly concentrated — with the inner and western Sydney LGAs of CBD, Parramatta, and Bankstown consistently recording the highest incident counts. Eight predictive models were benchmarked across three families; the traditional time-series model ARIMA achieved the lowest MAE (162.21), performing comparably to LSTM (163.92) and outperforming all ensemble learning methods. This result suggests that, for daily near-miss count prediction with the current data volume, parsimonious models capture the dominant temporal structure as effectively as more complex architectures.

\textbf{Policy implications:}

These findings carry direct relevance for transport planners and local government authorities. The consistent identification of CBD, Parramatta, and Bankstown as high-risk LGAs (driven by high traffic density and complex road geometry) provides a spatial basis for targeted road safety interventions. Recommended actions include prioritised deployment of variable message signs and speed enforcement during peak risk periods (weekday morning and afternoon peaks, particularly Wednesdays and Fridays), infrastructure review at high-incident corridors on trunk and primary roads, and integration of near-miss telemetry data into existing road safety auditing frameworks. The predictive capacity of the framework creates an operational window for pre-emptive resource deployment by traffic management centres, directly contributing to Australia's national goal of reducing road trauma. It should be noted, however, that these LGA rankings are derived from data collected during 2020--2021, a period affected by COVID-19 lockdowns that substantially suppressed traffic volumes across New South Wales (see Limitations). Since lockdown-related suppression affected metropolitan LGAs broadly rather than selectively, the \textit{relative} spatial ranking of high-risk areas is likely preserved even if absolute incident counts were reduced — a view supported by \citet{IEEEITSC25_Artur}, whose post-pandemic (2022) analysis of the same Compass IoT dataset identified near-miss hotspots concentrated in the same inner-city areas. Nevertheless, the specific spatial prioritisation should be re-validated against post-pandemic data before operational deployment of enforcement or infrastructure resources. The concentration of risk in Parramatta and Bankstown — LGAs with above-average socio-economic diversity — also raises an equity dimension that this paper does not address: future work should examine whether lower-income communities face disproportionate exposure to dangerous driving conditions, and how risk communications could be effectively targeted to reach drivers and residents in high-risk zones. This dimension aligns directly with the ATRF 2026 theme of integrating People, Place and Technology in transport planning.

\textbf{Limitations and future studies:}

The dataset used in this study covers April 2020 to December 2021, a period significantly affected by COVID-19 lockdowns in New South Wales. The suppression of incident counts during restriction periods reflects reduced traffic volumes rather than genuine improvements in driving behaviour, and may have influenced model training and performance. Future studies should isolate pandemic-affected periods or apply appropriate corrections before drawing policy-relevant conclusions.

While the dataset provides extensive geo-spatial coverage spanning over 700,000 connected vehicles from 64 manufacturers, the volume of daily aggregated observations remains insufficient for training data-hungry deep learning architectures, explaining the competitive performance of classical time-series models. The rolling origin evaluation window covers the final 35 days of each LGA's series, which, while chronologically rigorous and comprising 35 sequential forecast origins, is a relatively short absolute horizon; replication across longer held-out periods as new data accrues would further strengthen confidence in the reported model rankings. Future studies should explore richer feature sets including road geometry, speed limits, and weather conditions through multi-source data fusion. Longer time horizons and multi-step forecasting windows would further test the generalisability of the comparative model performance reported here. Additionally, the LGA-level rankings presented in \cref{Data_Mining_and_Analytics} are based on absolute incident counts rather than exposure-normalised rates; as shown by \citet{IEEEITSC25_Artur}, the near-miss signal captures latent risk beyond traffic volume, but future work should incorporate vehicle-kilometres travelled per LGA to further disentangle behavioural risk from exposure effects.

\textbf{Towards proactive road intelligence:}

The present work constitutes a foundational step towards the next generation of \textit{proactive road intelligence} systems. Building on the LGA-level risk maps and predictive models established here, our ongoing work is developing a real-time near-miss prediction framework that leverages continuously streaming connected vehicle telemetry to construct live situational awareness of dangerous driving conditions. Unlike the daily aggregate predictions reported in this paper, the target capability is to anticipate dangerous locations \textit{before} a driver reaches them — issuing warnings or informing dynamic routing in near real-time. Achieving this requires a finer spatial resolution than the LGA level used here, with modelling at street-segment, intersection, or grid-cell level, and the integration of live traffic state, signal timing, and infrastructure geometry as contextual features. This ongoing research will be reported in forthcoming publications. Taken together, this work and its extensions represent a shift from reactive crash reporting to predictive, data-driven road safety — where connected vehicle intelligence enables transport agencies to prevent crashes before they happen rather than respond after the fact.

% ============================================================
\section*{Acknowledgements}
% ============================================================

 The authors of this work are highly grateful for the data licence provided by Compass IoT (Angus McDonald and David Lillo-Trynes). This research is funded by the UTS Jenny Edwards Fellowship granted to Assoc. Prof. Adriana-Simona Mihaita for conducting research on connected vehicles and risky driver behaviour.

% \bibliography{} automatically generates the "References" section heading
% via natbib's \bibsection — do NOT add an explicit \section*{References} here.
\bibliographystyle{agsm}
\bibliography{ATRF}

\end{document}